\documentclass[lettersize,journal]{IEEEtran}
\usepackage{amsmath,amsfonts}
\usepackage{algorithmic}
\usepackage{algorithm}
\usepackage{array}
\usepackage[caption=false,font=normalsize,labelfont=sf,textfont=sf]{subfig}
\usepackage{textcomp}
\usepackage{stfloats}
\usepackage{url}
\usepackage{verbatim}
\usepackage{graphicx}
\usepackage{cite}
\usepackage{multirow}
\usepackage{multicol}
\usepackage{xcolor}
\usepackage{tikz}
\usepackage{enumitem}
\usepackage{makecell}

\begin{document}

\title{Simplifying Data-Driven Modeling of the Volume-Flow-Pressure Relationship in Hydraulic Soft Robotic Actuators}

\author{Sang-Yoep Lee,~\IEEEmembership{Member,~IEEE,}, Leonardo Zamora Ya\~nez, Jacob Rogatinsky, Vi T. Vo, Tanvi Shingade, and Tommaso Ranzani,~\IEEEmembership{Member,~IEEE,}
\thanks{
This work was supported by the National Institute of Biomedical Imaging and Bioengineering of the NIH, award R01EB035574.
The content is solely the responsibility of the authors and does not necessarily represent the official views of the NIH.
\emph{(Corresponding author: Tommaso Ranzani.)}}%

\thanks{Sang-Yoep Lee, Leonardo Zamora Ya\~nez, Jacob Rogatinsky, Vi T. Vo, and Tanvi Shingade are with the Department of Mechanical Engineering, Boston University, Boston, MA 02215 USA (e-mail: slee6@bu.edu; zamora@bu.edu; jrogat@bu.edu; vitvo@bu.edu; tshing26@bu.edu).}%

\thanks{Tommaso Ranzani is with the Departments of Mechanical Engineering, Biomedical Engineering, and the Materials Science and Engineering Division, Boston University, Boston, MA 02215 USA (e-mail: tranzani@bu.edu).}%
}

\IEEEpubid{}

\maketitle
\begin{abstract}
Soft robotic systems are known for their flexibility and adaptability, but traditional physics-based models struggle to capture their complex, nonlinear behaviors.
This study explores a data-driven approach to modeling the volume-flow-pressure relationship in hydraulic soft actuators, focusing on low-complexity models with high accuracy. 
We perform regression analysis on a stacked balloon actuator system using exponential, polynomial, and neural network models with or without autoregressive inputs. 
The results demonstrate that simpler models, particularly multivariate polynomials, effectively predict pressure dynamics with fewer parameters. 
This research offers a practical solution for real-time soft robotics applications, balancing model complexity and computational efficiency. 
Moreover, the approach may benefit various techniques that require explicit analytical models.
\end{abstract}

 
\section{Introduction}
\IEEEPARstart{S}{oft} robotic technologies have gained increasing interest over the past decade, leading to more adaptable, safe, and resilient robotic devices~\cite{rus2015design}.
Recent studies provide compelling evidence that a deeper understanding and integration of high-order dynamics and internal state variables in soft robots can significantly broaden their functionality and range of applications~\cite{della2020model}, enabling a shift from kinematics and quasi-statics to kinetics and dynamics~\cite{haggerty2023control}. 
However, these aspects have often been neglected due to challenges in observing and utilizing them. 
Unlike rigid robotics, where a standard equation of motion exists, there is no gold-standard mathematical framework for soft robotic behavior, which typically involves hysteresis, elastic deformation, and fluid dynamics~\cite{armanini2023soft}.

Sensor integration is essential not only for accurate modeling and control but also for robust, real‑time dynamic operation of soft robots~\cite{wang2018toward}. 
However, integrating sensors in soft robots is challenging as they undergo large, continuous, and nonlinear deformations that require a theoretically infinite number of measurement points and hinder reliable sensor attachment~\cite{lipson2014challenges}.
Even with high‑resolution measurements from a limited number of sensors, their dynamic properties remain inherently undersampled.
Thus, it can be critical to augment the observability of a soft robot's dynamics by leveraging intrinsic measurements from its fundamental subsystems such as actuation unit, beyond solely relying integrated sensors.

Among the actuation techniques for soft robots, fluidic actuation has been widely employed because of their large strokes, fast response, and high force transmission efficiency~\cite{polygerinos2017soft}. 
There are two main classes of fluidic actuators: elastomeric and sheet‑based designs.
Elastomeric fluid actuators achieve motion through stretching of their constituent materials, following the work input from a fluidic source that often reaches to high actuation pressures (i.e., hundreds of $kPa$)~\cite{cianchetti2014soft}. 
Sheet-based fluidic actuators, by contrast, are fabricated by selectively adhering thin, less-extensible films, and achieve motion via the geometric unfolding of these layers at much lower fluidic pressures (i.e., tens of $kPa$)~\cite{niiyama2015pouch}. 

A prominent example of the sheet‑based approach is the stacked balloon actuator (SBA)~\cite{SBA1}.
SBAs are built from thin sheet materials so they occupy almost no volume when deflated but produce large, preprogrammed strokes when inflated.
By integrating multiple SBA stacks in parallel, multidirectional bending is achievable for applications ranging from beating‑heart surgery~\cite{SBAmed2}, to dexterous manipulation~\cite{SBAlike2}, to reconfigurable robot locomotion~\cite{SBA3}.

To describe the frequency-dependent behavior of a single soft fluidic actuator, Mosadegh et al. introduced an analytic framework based on volume--pressure curves (V--P curves; denoted \textit{PV curves} in the original article)~\cite{mosadegh2014pneumatic}.
The V--P curve shows a unique profile for each actuator design, actuation condition, and environment.
Due to pressure uniformity in the actuation fluid, the pressure can be measured remotely (e.g., in the fluidic supply line) without interfering the soft actuator’s behavior.
Since the pressure reflects changes in stroke and velocity, it can be used as a primary indicator of the actuator’s dynamic state.

In conventional rigid-body robotics, system dynamics and mechanical impedance are analyzed by quantifying how twist and its time derivatives correlate with wrench.
In fluidically actuated soft systems, an analogous framework can be built on fluid volume and its time derivatives to capture motion, and on pressure to capture force. 
This approach not only leverages established analytical tools in robotics but also establishes a foundation for the effective control and performance optimization of soft robots.
However, the nonlinear nature of the soft material combined with fluid dynamics complicates establishing such relationship using simple physics-based laws.
Moreover, factors such as fabrication defects or air pockets in hydraulic channels introduce actuator‑specific variability, which underscores the need for adaptive, data‑driven modeling.

\begin{figure}[!t]
\centering
\includegraphics[width=0.48\textwidth]{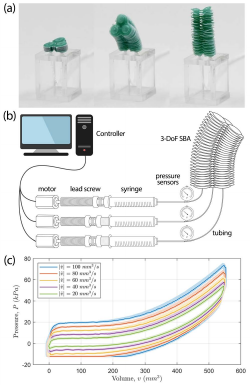}
\caption{
\textbf{Overview of the experimental setup with a 3-DoF SBA.}
(a)~Fully deflated, partially inflated, and fully inflated 3-DoF SBA used in experiments.
(b)~Schematic overview of the actuation system for 3-DoF SBA. 
(c)~V--P plot for a benchmark SBA across multiple inflation and deflation cycles at different $|\dot{v}|$. 
The shaded areas cover the range of actual trajectories for each flow rate, while the solid lines represent the average trajectory.
}
\label{fig_1}
\end{figure}

In this study, we introduce a modeling approach to achieve the (fluidic-)actuation-to-pressure relationship in an inflatable soft actuator that balances prediction accuracy and model complexity. 
Taking a three-degree-of-freedom (3-DoF) SBA (Fig.~\ref{fig_1}(a)) based on~\cite{SBAmed2} as a benchmark platform to collect datasets, three types of models, physics-inspired exponential, neural network, and multivariate polynomial models, were proposed and navigated.
After identifying the model that offers the optimal balance between accuracy and complexity, we used it to tackle three key challenges associated with inflatable soft robots: dynamic characterization of SBAs to detect the presence of air pockets, 
differentiation of SBAs varying in size and design through tailored characterization, 
and proprioceptive estimation of external forces.
Along with the potential to expand soft robotic analysis toward dynamics using the proposed modeling techniques, we also share important findings used throughout the study, such as the fact that fluid volume and flow rate (without acceleration) dominantly governs the pressure change and that the prediction performance can be improved by involving previous step terms.

\section{Modeling Volume-Flow-Pressure Relation}
To investigate the volume-flow-pressure relationship, a 3-DoF SBA based on the design introduced in~\cite{SBAmed2} was selected as the benchmark SBA.
The SBA actuation system and experimental setup for the 3-DoF SBA are shown in Fig.~\ref{fig_1}(b). 
For actuating the three chambers of the SBA, we used three actuation units, one per chamber.
Each unit consisted of a stepper motor (NEMA 11, STEPPERONLINE), lead screw, medical-grade syringe (7649-01, Hamilton), and pressure sensor (P-7100, Nidec Components).
Directly attached to each motor shaft, the lead screw stages were connected to the plunger of the syringe with a 2~$mm$ diameter tube. 
The syringe volumes were controlled with a precision of 0.04~$mm^3$ per step, following the motors' specifications.
Hydraulic actuation was used throughout the study.
The data were collected by injecting and withdrawing water from 0~$mm^3$ to 550~$mm^3$ while varying the volume flow rate up to 100~$mm^3/s$. 
The data points were sampled at 25 $Hz$.
Minimum 20 inflation/deflation cycles were recorded for each experimental condition (e.g., inflation volume, flow rate, and acceleration) to account for hysteresis in the V--P profile.

In Fig.~\ref{fig_1}(c), volume-pressure profiles for five flow rates ($\dot{v}(t)$) from 20~$mm^3/s$ to 100~$mm^3/s$ are shown.
To evaluate the extent to which $P(t)$ can be explained by $v(t)$ and its time derivatives, we first calculated the correlations between $P(t)$ and $v(t)$, $P(t)$ and $\dot{v}(t)$, and $P(t)$ and $\ddot{v}(t)$.
The three correlations were: $corr(P, v) = 0.8029$, $corr(P, \dot{v}) = 0.4496$, and $corr(P, \ddot{v}) = 0.0965$.
Since the tubing used in hydraulic actuation involves fluidic friction and viscosity, these results indicate that the dynamics are primarily governed by $v(t)$ and $\dot{v}(t)$, with little contribution from $\ddot{v}(t)$. 

Principal Component Analysis (PCA) was also performed with the dataset. 
The dataset was defined as:

\begin{equation}
X =    
    \begin{bmatrix}
        v(t_1) & \dot{v}(t_1) & \ddot{v}(t_1) & P(t_1) \\
        v(t_2) & \dot{v}(t_2) & \ddot{v}(t_2)  & P(t_2) \\
        \vdots & \vdots & \vdots & \vdots \\
        v(t_N) & \dot{v}(t_N) & \ddot{v}(t_N)  & P(t_N) 
    \end{bmatrix}
\end{equation}
where $X \in \mathbb{R}^{N \times 4}$ consisting of columns $[v, \dot{v}, \ddot{v}, P]$ over $N$ time points, with each row representing the corresponding values of volume $v(t)$, volumetric flow rate $\dot{v}(t)$, volumetric acceleration $\ddot{v}(t)$, and pressure $P(t)$ at a specific $t=t_i$.
PCA was performed on the normalized data matrix $X' = (X - \mu)/{\sigma}$, where $\mu$ and $\sigma$ as the mean and standard deviation of each column of $X$, respectively. 
The eigenvectors $V$ and normalized eigenvalues $\vec{\lambda}_{\mathrm{norm}}$ were computed according to $\Bigl(\frac{1}{N-1}X'^TX'\Bigr)V = \Lambda V$ and $\vec\lambda_{norm}=\frac{\vec\lambda}{tr(\Lambda)}$, where $N$ is the number of observations, $\Lambda = \mathrm{diag}(\vec{\lambda})$ is the matrix of eigenvalues, and $V$ is the matrix of eigenvectors.

\begin{table}[!t]
\caption{Preliminary Analysis: PCA Results}
\centering
\begin{tabular}{cc|cccc}
 \multicolumn{2}{c|}{} & PC1 & PC2 & PC3 & PC4 \\
 \hline
 \multirow{4}*{$V$} & $v(t)$ & 0.61 & -0.31 & -0.38 & -0.62\\
 & $\dot{v}(t)$ & 0.34 & 0.70 & -0.53 & -0.35\\
 & $\ddot{v}(t)$ & -0.13 & 0.64 & 0.75 & 0.02\\
 & $P(t)$ & 0.70 & 0.05 & 0.06 & 0.71 \\
\hline
\multicolumn{2}{c|}{$\vec\lambda_{norm}$} & 0.48 & 0.26 & 0.24 & 0.02 \\
\end{tabular}
\label{PCA_results}
\end{table}

\begin{table*}[!b]
\caption{Pressure Estimation Models}
\centering
\begin{tabular}{c|l}
\hline\hline
 Models & Equations \\
 \hline
\multirow{2}{*}{\makecell{Physics-inspired\\exponential model}} & \(\displaystyle \hat{P}_{exp}(t) = \sum_{i=1}^{k}\alpha_i\cdot exp(\beta_iv(t))+\gamma\dot{v}(t)+\delta\), \\
& $k$: the number of exponential terms; $\alpha_i, \beta_i, \gamma, \delta$: model parameters determined from data.\\\hline
\multirow{3}{*}{\makecell{Neural network model}} & \(\displaystyle \hat{P}_{\mathrm{NN}}(t) = f_{\mathrm{NN}}\Bigl(v(t),\dot{v}(t);\ \theta_{\mathrm{NN}} \Bigr)\),\\
& $f_{\mathrm{NN}}$: a fully connected network with two inputs $\{v(t),\dot{v}(t)\}$ and one scalar output $\hat{P}_{\mathrm{NN}}(t)$;\\
& $\theta_{\mathrm{NN}}$: the parameter vector including trained weights and biases.\\\hline
\multirow{2}{*}{\makecell{Autoregressive\\neural network model}} & \(\displaystyle \hat{P}_{\mathrm{NN\_AR}}(t) = f_{\mathrm{NN\_AR}}\Bigl(\{v(t-\Delta t_i), \dot{v}(t-\Delta t_i)\}_{i=0\dots p};\ \theta_{\mathrm{NN\_AR}}\Bigr)\)\\
& $f_{\mathrm{NN\_AR}}$ extends $f_{\mathrm{NN}}$ to include $p$ previous time steps ${v(t-\Delta t_i),\dot v(t-\Delta t_i)}_{i=0...p}$, with $\Delta t_0=0$.\\\hline
\multirow{2}{*}{\makecell{Multivariate\\polynomial model}} & \(\displaystyle \hat{P}_{poly}(t) = \sum_{i=0}^{n}\sum_{j=0}^m\alpha_{i,j}\bigl(v(t)\bigr)^i\bigl(\dot{v}(t)\bigr)^j\)\\
& $n, m$: the highest degrees of $v$ and $\dot{v}$, respectively; $\alpha_{i,j}$: a parameter to be determined from experimental data.\\\hline
\multirow{2}{*}{\makecell{Autoregressive multivariate\\polynomial model}} & \(\displaystyle \hat{P}_{poly\_AR}(t) = \sum_{i=0}^{n}\sum_{j=0}^m\alpha_{i,j}\bigl(v(t)\bigr)^i\bigl(\dot{v}(t)\bigr)^j 
+\sum_{k=1}^{p}\Bigl(\beta_{k,1} v(t-\Delta t_k)+\beta_{k,2}\dot{v}(t-\Delta t_k)\Bigr)\)\\
& $\alpha_{i,j}, \beta_{k,1}, \beta_{k,2}$: parameters to be determined from experimental data.\\\hline
\end{tabular}
\label{Eqss}
\end{table*}

Table~\ref{PCA_results} summarizes the PCA results.
Among the four principal components, PC1 and PC4 include relatively large weights for $P(t)$  (0.70 and 0.71, respectively) compared to PC2 and PC3 (0.05 and 0.06). 
In both PC1 and PC4, $v(t)$ and $\dot{v}(t)$ shows larger weights than $\ddot{v}(t)$ (0.61~and~0.34~$>$~0.13 in PC1; 0.62~and~0.35~$>$~0.02 in PC4).
Considering the normalization applied to the dataset, these relative magnitudes suggest that the contributions of $v(t)$ and $\dot{v}(t)$ dominate those of $\ddot{v}(t)$, as further supported by the calculated correlations.

The strong correlation of $P(t)$ with $v(t)$ and $\dot{v}(t)$ motivates the inclusion of higher-degree cross terms, such as $v(t) \cdot \dot{v}(t)$, to better capture the underlying nonlinear dynamics since the dominant components of $P(t)$ arose from combinations of $v(t)$ and $\dot{v}(t)$.
This insight justify the need to investigate more complex models to fully capture the system's dynamics.

Based on these PCA results, we modeled $P(t)$ using $v(t)$ and $\dot{v}(t)$, excluding $\ddot{v}(t)$.
The mathematical form of the models used in the following sections are summarized in Table~\ref{Eqss}.

\subsection{Physics-Inspired Exponential Model}
From a physical standpoint, two dominant behaviors govern the pressure inside a soft robotic system. 
First, the hyper-elastic properties of the material drive a nonlinear expansion as the internal volume changes~\cite{neohook2}.
%
Second, the fluid flow through the tubing can be assumed to be governed by Hagen–Poiseuille’s law, indicating that the flow rate influences the fluid pressure~\cite{hagen1}.
We model these two effects as additive contributions: one term addressing the chamber’s elastic response, and another accounting for the flow resistance.

Numerous analytic basis functions can serve to approximate the complex and nonlinear nature of soft robotic materials. 
In this section, we focused on exponential terms in volume, deferring a polynomial model description to Section~\ref{s2-3}. 
The sum of exponential terms was used to capture the elastic expansion of the SBA chamber as a representation of the hyper-elastic relationship, while the affine term in $\dot{v}$ addresses the Hagen–Poiseuille’s law. 

%
By analyzing the impact of numbers of exponential terms, this model was used to assess the best achievable fit accuracy without volume-flow interaction terms.

\subsection{Neural Network Model}
Neural network models were introduced to assess the upper bound of predictive performance as model complexity increases without forcing an explicit analytical interaction between $v(t)$ and $\dot{v}(t)$.
While a large size neural network may guarantee high accuracy, we kept the network’s parameter count in the same order of magnitude as that of the other models for comparison.
The Rectified Linear Unit (ReLU) function was used as the activation function for all neural network models due to its advantages, such as computational efficiency, stability, and the avoidance of vanishing gradients, compared to other activation functions like sigmoid or hyperbolic tangent.

%
Neural network models with autoregressive inputs were also investigated to check if estimation performance could be improved when incorporating short-term memory of state changes while dealing with possible temporal factors (e.g., delays).
%

\subsection{Multivariate Polynomial Model} \label{s2-3}
Based on insights from the results of physics-inspired exponential and neural network models, multivariate polynomial models were introduced to explore the effect of higher-degree cross-terms of $v$ and $\dot{v}(t)$, while retaining an explicit analytical form.
Compared to the physics-inspired exponential model, the way we capture hyper-elastic behavior was changed from using exponential terms to using polynomial terms of $v$ and $\dot{v}$.

%
Multivariate polynomial models with autoregressive terms were also evaluated.
Rather than considering all possible combinations of $v$, $\dot{v}$, and their previous terms, only linear autoregressive terms were included.
This choice kept the number of terms below 20 while still accounting for temporal effects from the previous terms.

\subsection{Model Evaluation Criteria}
To assess both the predictive accuracy and the complexity of the candidate models, we employed the following four metrics:

\begin{itemize}
    \item \textbf{Root Mean Squared Error (RMSE)}.
    This measures the typical magnitude of the prediction error:
    \begin{equation}
    \mathrm{RMSE} = \sqrt{\frac{1}{N}\sum_{i=1}^{N}\Bigl(P(t_i)-\hat{P}(t_i)\Bigr)^2},
    \end{equation}
    where $\hat{P}$ denotes the predicted value, and $N$ is the total number of data points. 
    A lower RMSE indicates smaller residuals on average.
    \item \textbf{Adjusted \boldmath{$R^2$}}.
    Unlike the standard $R^2$ value, the adjusted $R^2$ penalizes model complexity. 
    For a model with $k$ predictors fit to $N$ data samples, the adjusted $R^2$ is defined as:
    \begin{equation} R^2_{\mathrm{adj}} = 1 - \frac{\sum_{i=1}^N(P(t_i)-\hat{P}(t_i))^2}{\sum_{i=1}^N(P(t_i)-\bar{P})^2} \cdot \frac{N-1}{N - k - 1},
    \end{equation}    
    where $\bar{P} = \sum_{i=1}^N P(t_i)/N$.
    The adjusted $R^2$ increases only if a newly added predictor improves the model more than would be expected by chance.

    \item \textbf{Corrected Akaike Information Criterion (AICc)}.
    As a model selection tool, the Akaike information criterion (AIC) estimates the prediction error to compare the relative quality of candidate models for a given dataset~\cite{hurvich1993corrected}
    It balances the goodness of fit against model simplicity, addressing both overfitting and underfitting. 
    The standard form can be written as $ \mathrm{AIC} = 2\nu - 2\ln(\mathcal{L})$, where $\nu$ is the number of estimated parameters and $\ln(\mathcal{L})$ is the log-likelihood of the model.
    Under the assumption of Gaussian regression, the log-likelihood term $\ln(\mathcal{L})$ is approximated by the following equation~\cite{gaussian1}:
    \begin{equation}
    \ln(\mathcal{L}) \approx -\frac{N}{2}\Bigl[\ln\frac{2\pi}{N}\sum_{i=1}^N\Bigl(P(t_i)-\hat{P}(t_i)\Bigr)^2+1\Bigr].
    \label{loglike}
    \end{equation}
    In this work, we employed the AICc to more strongly penalize model complexity~\cite{hurvich1993corrected}.
    For finite sample sizes, the AICc introduces an additional penalty term:
    \begin{equation}
    \mathrm{AICc} = \mathrm{AIC} + \frac{2\nu(\nu+1)}{N-\nu-1}.
    \end{equation}
    Lower values of AICc indicate a better trade-off between fit quality and parameter count, especially when $N$ is not significantly larger than $\nu$. 
    Aside from AICc, there have been several studies that utilized the Bayesian information criterion (BIC), given by $\mathrm{BIC} = \nu\ln(N) - 2\ln(\mathcal{L})$, for model optimization.
    BIC generally emphasizes more on larger values of $N$ than AICc.
    However, for our datasets, the absolute values of AICc and BIC tended to be nearly identical, while AICc showed higher sensitivity to $\nu$. 
    This difference can be shown with their partial derivatives:
    \begin{equation}        
        \frac{\partial \mathrm{AICc}}{\partial \nu} = 2+\frac{(4\nu+2)(n-\nu-1)+2\nu(\nu+1)}{(n-\nu-1)^2}
    \end{equation}
    and
    \begin{equation}        
        \frac{\partial \mathrm{BIC}}{\partial \nu} = \ln(N).
    \end{equation}
    Thus, AICc was more appropriate in our case.
    \item \textbf{Time Complexity}.
    For real-time state estimation and control of systems, optimizing computation efficiency is important.
    Given the number of data points $N$, the number of parameters $\nu$, the data matrix $X\in \mathbb{R}^{N \times \nu}$, and the regression vector $y\in\mathbb{R}^{N}$, the time complexity for all the physics-based exponential models and multivariate polynomial models is estimated as $O(N\cdot\nu^2)$ due to computation of least squares parameters $(X^TX)^{-1}X^Ty$. 
    On the other hand, the time complexity for the neural network model can be written as $O(epochs\cdot N\cdot \nu)$ where the number of parameters $\nu$ equals the number of network weights, which grows with neurons and layers.

\end{itemize}

\subsection{Model Optimization}
For each model, the optimal parameters were determined by minimizing a joint cost function that linearly combines the evaluation criteria introduced in the previous section:
\begin{equation}
\label{eq_optimization}
\theta^* = \underset{\theta}{\mathrm{arg\,min}}\Bigl[w_1\mathrm{RMSE}+w_2(1-R^2_{\mathrm{adj}})+w_3\mathrm{AICc}\Bigr]
\end{equation}
where $w_1, w_2, w_3$ are weights balancing the relative importance of predictive error and model complexity. 
Note that one's complement of the term $R^2_{\mathrm{adj}}$, $1-R^2_{\mathrm{adj}}$ was used for minimization consistency. 
In many cases, reducing RMSE simultaneously lowered $(1-R^2_{\mathrm{adj}})$ and AICc, and encouraged the removal of unnecessary components (e.g., dropping negligible terms or neurons) when possible.
Throughout this procedure, the model reduction was performed without harming the model's general form (i.e. the number of exponential terms or highest exponents of $v(t)$ and $\dot{v}$).

\begin{table}[!t]
\caption{Summary of Regression Performances}
\label{exp_results}
\centering
\begin{tabular}{l|cccc}
 \hline\hline
 \multicolumn{5}{c}{Physics-inspired exponential model} \\
\hline
 $k$ & $\nu$ & RMSE & $R^2_{\mathrm{adj}}$ & AICc \\
 \hline
 1 & 4 & 1.2986 & 0.9934 & 2.7573e+5 \\
 2 & 6 & 1.1100 & 0.9952 & 2.4999e+5 \\
 3$^*$ & 8 & 1.0998 & 0.9952 & 2.4849e+5 \\
 4 & 10 & 1.0998 & 0.9952 & 2.4848e+5 \\
 5 & 12 & 1.0997 & 0.9952 & 2.4847e+5 \\
 \hline\hline
 \multicolumn{5}{c}{Neural network model} \\ \hline
  $d$ & $\nu$ & RMSE & $R^2_{\mathrm{adj}}$ & AICc \\ \hline
 1 & 5 & 4.4011 & 0.9239 & 4.7604e+5 \\
 2 & 9 & 0.7067 & 0.9980 & 1.7591e+5 \\
 & \multicolumn{4}{c}{\dots}  \\
 7 & 29 & 0.4292 & 0.9993 & 9.4124e+4 \\
 8$^*$ & 33 & 0.3213 & 0.9996 & 4.6604e+4 \\
 9 & 37 & 0.3150 & 0.9996 & 4.3371e+4 \\
 & \multicolumn{4}{c}{\dots}  \\
 14 & 57 & 0.2947 & 0.9997 & 3.2574e+4 \\
 15 & 61 & 0.2945 & 0.9997 & 3.2335e+4 \\
 \hline\hline
 \multicolumn{5}{c}{Autoregressive neural network model} \\ \hline
 ($p$, $d$) & $\nu$ & RMSE & $R^2_{\mathrm{adj}}$ & AICc \\
 \hline
 (0, 8) & 33 & 0.3319 & 0.9995 & 2.6367e+4 \\
 (1, 8) & 41 & 0.3324 & 0.9995 & 2.6649e+4 \\ 
 (2, 8) & 57 & 0.3302 & 0.9995 & 2.6027e+4 \\
 (3, 8) & 73 & 0.3304 & 0.9995 & 2.6020e+4 \\
 (4, 8) & 89 & 0.3330 & 0.9995 & 2.6616e+4 \\
 (5, 8)$^*$ & 105 & 0.3321 & 0.9995 & 2.6321e+4 \\
 \hline\hline
 \multicolumn{5}{c}{Multivariate polynomial model} \\ \hline
  ($n$, $m$) & $\nu$ & RMSE & $R^2_{\mathrm{adj}}$ & AICc \\
 \hline
 (2, 1) & 6 & 1.4291 & 0.9920 & 2.9146e+5 \\
 (2, 2) & 9 & 1.2165 & 0.9942 & 2.6503e+5 \\
 (2, 3) & 12 & 1.4097 & 0.9922 & 2.8922e+5 \\
 (3, 1) & 8 & 1.0644 & 0.9955 & 2.4312e+5 \\
 (3, 2)$^*$ & 12 & 0.6884 & 0.9981 & 1.7160e+5 \\
 (3, 3) & 16 & 1.1861 & 0.9945 & 2.6090e+5 \\
 (4, 1) & 10 & 2.6945 & 0.9715 & 3.9554e+5 \\
 (4, 2) & 15 & 2.6926 & 0.9715 & 3.9542e+5 \\
 \hline\hline
 \multicolumn{5}{c}{Autoregressive multivariate polynomial model} \\ \hline
 ($p$, $n$, $m$) & $\nu$ & RMSE & $R^2_{\mathrm{adj}}$ & AICc \\
 \hline
 (0, 3, 2) & 12 & 0.4603 & 0.9990 & 5.4029e+4 \\
 (1, 3, 2) & 14 & 0.4521 & 0.9991 & 5.2339e+4 \\ 
 (2, 3, 2) & 16 & 0.4450 & 0.9991 & 5.0829e+4 \\
 (3, 3, 2) & 18 & 0.4388 & 0.9991 & 4.9480e+4 \\
 (4, 3, 2) & 20 & 0.4332 & 0.9992 & 4.8242e+4 \\
 (5, 3, 2)$^*$ & 22 & 0.4280 & 0.9992 & 4.7063e+4 \\
 \hline
\end{tabular}
\begin{itemize}
    \item[] \footnotesize{$k$: number of exponential terms; $\nu$: degree(s) of freedom or number of parameters; $d$: depth of fully connected neural network; $p$: order of autoregressive model; $n$: maximum exponent of volume; $m$: maximum exponent of volume change rate.}
    \item[] \footnotesize{$^a$ Models with $^*$ are pseudo-optimal sets within each model type.}
    \item[] \footnotesize{$^b$ AICc scores can be compared within each model type.}    
\end{itemize}
\end{table}

\section{Model Evaluation}
In this section, we evaluate and discuss the fitting performance of each model.
RMSE was the primary goodness-of-fit metric, and AICc values were plotted together to show relative model parsimony within each model category for visualization.
The $R^2_{\mathrm{adj}}$ values were all above 0.99 with the variance less than 0.01, so we omit them from plots for clarity.
All the detailed evaluation criteria scores are summarized in Table~\ref{exp_results}.

\begin{figure}[!t]
\centering
\includegraphics[width=0.48\textwidth]{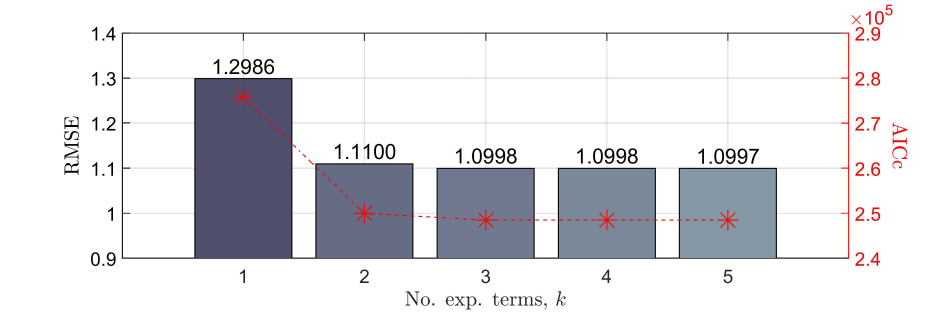}
\caption{
\textbf{Regression performances using a physics-inspired exponential model.}
The bar plot shows the RMSE values, while the red stars indicate the corresponding AICc values for varying numbers of exponential terms ($k$).
}
\label{fig_2}
\end{figure}

\subsection{Physics-Inspired Exponential Model}
Fig.~\ref{fig_2} shows results for physics-inspired exponential models with up to five exponential terms ($k=1,\dots,5$).
The estimation performance showed only a minor improvement ($\Delta$RMSE $<$ 0.01) beyond one exponential term. 
Including more than five exponential terms did not lead to significant improvement in performance ($\Delta$RMSE $<$ 0.0001).
AICc values followed the trend of RMSE values, which may result from the increase in model complexity being canceled out by the decrease in error.
Using three exponential terms showed the best balance of accuracy and simplicity (eight parameters, RMSE of 1.0998, $R^2_{\mathrm{adj}}$ of 0.9952, and AICc of 2.4849e+5).

Since the term $\dot{v}(t)$ was constrained to an affine form, increasing the number of exponential terms only increased the complexity in the $v(t)$ component of the model.
This result indicated that a more intricate coupling of $v(t)$ and $\dot{v}(t)$ is needed, as hinted by the PCA study.

\subsection{Neural Network Model}
\begin{figure}[!t]
\centering
\includegraphics[width=0.48\textwidth]{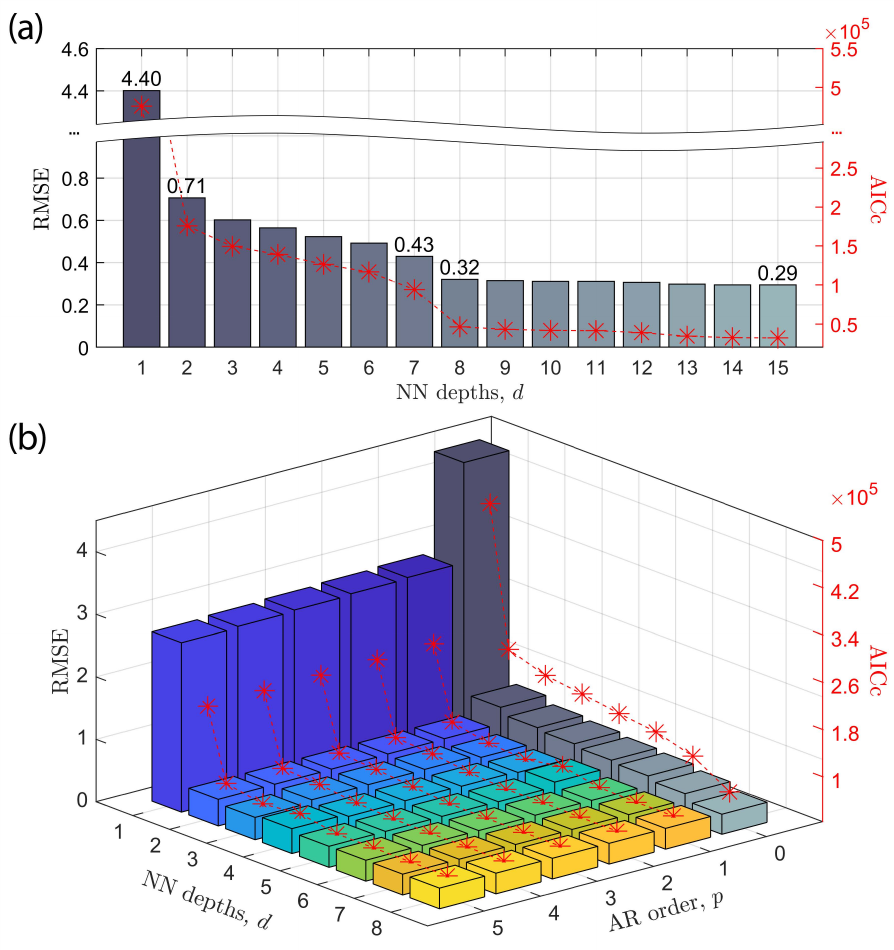}
\caption{
\textbf{Regression performances using neural network models.}
(a)~The bar plot shows the RMSE values, while the red stars indicate the corresponding AICc values for varying numbers of neural network depths, $d$, without autoregressive inputs.
(b)~The bar plot displays the RMSE values, while the red stars represent the corresponding AICc values for varying network depths, $d$, and autoregressive orders, $p$. 
The gray bars ($p = 0$) represent the performance of the neural network models without autoregressive inputs shown in (a).
}
\label{fig_3}
\end{figure}

The neural network models were evaluated by varying the number of layers up to 15 ($d=1,\dots,15$; Fig.~\ref{fig_3}(a)). 
A drastic decrease in the RMSE from 4.40 to 0.71 was observed after increasing the depth beyond one, as the network began to differentiate from a least-squares model of $v$ and $\dot{v}$ with a nonlinear activation function. 
The performance gradually improved until a depth of eight (RMSE of 0.71 with depth two and RMSE of 0.23 with depth eight), after which the performance saturated. 
Overall, the neural network models achieved better performance (lower RMSE and AICc) than the physics-inspired exponential models.
As for the physics-inspired exponential models, the AICc values followed a similar trend to the RMSE values, showing a drastic improvement after depth seven (AICc of 1.7591e+5 with depth two and AICc of 4.6604e+4 with depth eight).

Fig.~\ref{fig_3}(b) shows the results from the neural network models with autoregressive inputs. 
Models with up to eight layers and autoregressive order $p$ up to five were evaluated. 
Incorporating at least one previous time step significantly improved performance (reducing the RMSE up to 36\%, when increasing the AR order $p$ from 0 to 1)
Adding more than one autoregressive input did not lead to a substantial improvement (RMSE of 0.3324 with AR order one and 0.3321 with AR order five).

Two key insights were drawn from the results of the neural network models.
First, including at least one prior time step is important for pressure estimation.
Second, a regression model can be established without using more than 100 parameters (i.e., the simple neural network model with depth of 15 showed RMSE of 0.2945, $R^2_{\mathrm{adj}}$ of 0.9997, and AICc of 3.2335e+4, which was the best performance compared to all the other models).

\subsection{Multivariate Polynomial Model}
\begin{figure}[!t]
\centering
\includegraphics[width=0.48\textwidth]{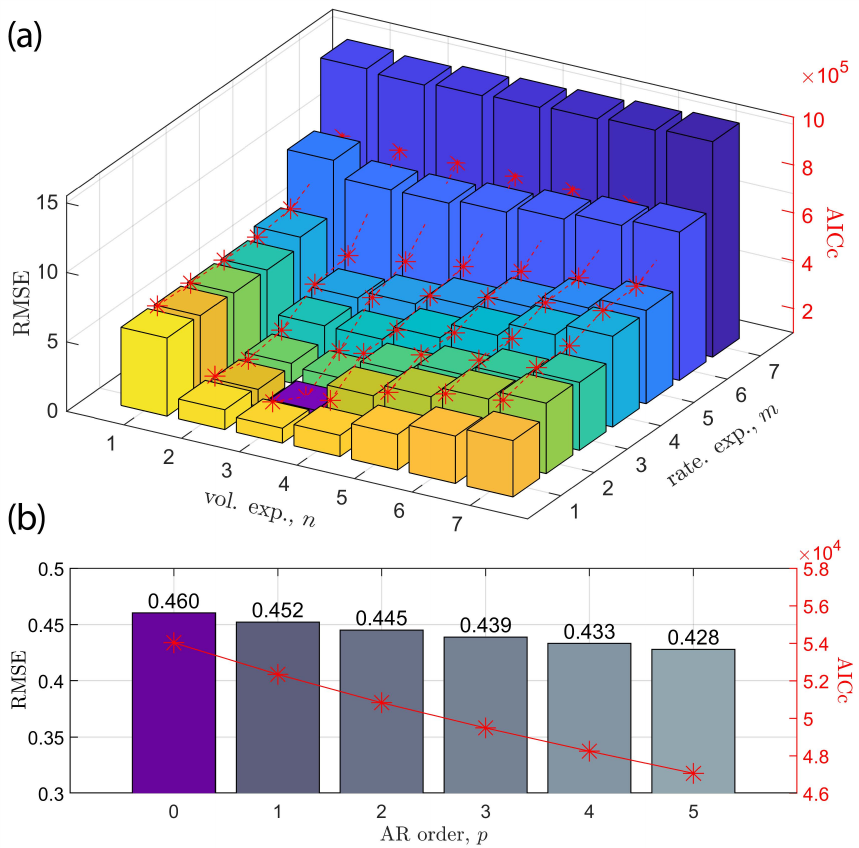}
\caption{
\textbf{Regression performances using multivariate polynomial models.}
(a)~The bar plot shows the RMSE values, while the red stars represent the corresponding AICc values for varying maximum exponents of volume, $n$, and maximum exponents of volume change rate, $m$.
The purple bar ($n = 3, m = 2$) indicates the best results among the adjacent models.
(b)~Regression performances using multivariate polynomial models with autoregressive inputs, where the maximum exponent of volume is three and the maximum exponent of volume change rate is two ($n = 3, m = 2$). 
The bar plot displays the RMSE values, while the red stars represent the corresponding AICc values for varying autoregressive orders, $p$. 
The purple bar ($p = 0$) shows the best performance multivariate polynomial model without autoregressive inputs.
}
\label{fig_4}
\end{figure}

The multivariate polynomial models were evaluated by varying both the maximum exponents for $v(t)$ and $\dot{v}(t)$ from one up to seven for each ($n=1,2,...,7$ and $m=1,2,...,7$), as shown in Fig.~\ref{fig_4}(a). 
The highest-degree term’s coefficients were constrained to be nonzero to prevent degenerate solutions (producing a reduced-degree model).
A drastic decrease in performance (increase in both RMSE and AICc values; around 138\% increase in RMSE) was observed after including terms with more than degree five.

Using a degree of three for $v(t)$ and a degree of two for $\dot{v}(t)$ ($n=3, m=2$) yielded significantly better performance than all the other polynomial models.
Notably, this model’s RMSE was 35\% to 74\% lower than that of models with nearby degree combinations ($n=2, m=1$ through $n=4, m=3$).
This result suggests that beyond a certain complexity, increasing the degree does not significantly improve the model’s ability to describe the system’s behavior. 
It also points to the potential for simpler models that still capture the essential dynamics of the system effectively.

To further investigate the identified optimal model ($n=3, m=2$), the performance was evaluated after excluding one or more terms based on their contributions to the model, which were assessed by the magnitude of the coefficients and p-values from a two-sided t-statistic, with the null hypothesis that each coefficient is zero.
The results are summarized in Table~\ref{poly_red}. 
In all cases, excluding a term degraded performance. 
The exclusion of terms was extended to adjacent models with different complexities, but no model outperformed the original model. 
Even when a parameter's magnitude is very small ($<10^{-10}$), it can be concluded that all the terms remained essential for accurately representing the dataset.

Taking insight from the neural network model with autoregressive inputs, multivariate polynomial models with linear autoregressive terms up to a degree of five were inspected for the optimal case ($n=3, m=2$). 
The results are shown in Fig.~\ref{fig_4}(b). 
Unlike the neural network models (Fig.~\ref{fig_3}(b)), here RMSE and AICc kept improving as the autoregressive order increased. 
Models with more autoregressive terms were not considered due to concerns about overfitting and a long initial transient.

\begin{table}[!t]
\caption{
Coefficients of the Multivariate Polynomial Model ($n=3, m=2$) and Regression Performances of Reduced Variants
}
\label{poly_red}
\centering
\begin{tabular}{cc|cc|cc}
 \hline\hline
 \multicolumn{6}{c}{Multivariate polynomial model: ($n=3, m=2$)} \\
\hline
term & coeff. & term & coeff. & term & coeff. \\ \hline
 1 & 0 & $\dot{v}$ & 1.7660e-1 & $\dot{v}^2$ & 2.3010e-4\\ 
 $v$ & 1.2695e-2 & $v\dot{v}$ & -1.3896e-4 & $v\dot{v}^2$ & -3.0150e-6\\
 $v^2$ & -8.0664e-5 & $v^2\dot{v}$ & 6.6527e-7 & $v^2\dot{v}^2$ & 1.3800e-8\\
 $v^3$ & 4.1269e-7 & $v^3\dot{v}$ & -7.3542e-10 & $v^3\dot{v}^2$ & -1.3773e-11 \\
 \hline\hline
 \multicolumn{6}{c}{Reduced multivariate polynomial models} \\ \hline
 \multicolumn{2}{c|}{excluded term(s)} & \multicolumn{1}{c}{$\nu$} & \multicolumn{1}{c}{RMSE} & \multicolumn{1}{c}{$R^2_{\mathrm{adj}}$} & \multicolumn{1}{c}{AICc}\\\hline
 \multicolumn{2}{c|}{$v\dot{v}$} & 11 & \multicolumn{1}{c}{0.6934} & 0.9981 & 1.7279e5 \\
 \multicolumn{2}{c|}{$v^3\dot{v}$} & 11 & \multicolumn{1}{c}{0.6943} & 0.9981 & 1.7301e5 \\
 \multicolumn{2}{c|}{$v\dot{v}$, $v^2\dot{v}$} & 10 & \multicolumn{1}{c}{0.6971} & 0.9981 & 1.7367e5 \\ 
 \multicolumn{2}{c|}{$v\dot{v}$, $v^3\dot{v}$} & 10 & \multicolumn{1}{c}{0.6943} & 0.9981 & 1.7301e5 \\ 
 \multicolumn{2}{c|}{$v\dot{v}$, $v^3\dot{v}$} & 10 & \multicolumn{1}{c}{0.6943} & 0.9981 & 1.7301e5 \\ 
 \multicolumn{2}{c|}{$v^2\dot{v}$, $v^3\dot{v}$} & 10 & \multicolumn{1}{c}{0.6962} & 0.9981 & 1.7345e5 \\ \hline
\end{tabular}
\begin{itemize}    
    \item[] \footnotesize{
    The least-contributing terms among the 12 terms of the multivariate polynomial model ($n = 3, m = 2$) were selectively excluded based on the p-value of the t-statistic for a two-sided test with the null hypothesis that the coefficient is zero.
    }
\end{itemize}
\end{table}

\subsection{Comparative Performance Review}
Throughout our results, the number of parameters used in the physics-inspired exponential models and multivariate polynomial models was equal to or less than 22 ($\nu\leq22$).
On the other hand, the neural network models involved a larger number of parameters ($\nu\leq105$) and required up to 3,000 training epochs to converge.
The actual computational time for inspecting all the possible models within each model category (e.g., physics-inspired exponential model, neural network model, and multivariate polynomial model) matched the time complexity analysis in the previous section ($O(N\cdot\nu^2)$ for the physics-inspired exponential models and multivariate polynomial models; $O(epochs\cdot N\cdot \nu)$ for the neural network models).
The physics-inspired exponential models required 2.31 $s$ to optimize models up to 10 exponential terms, and multivariate polynomial models took 13.03 $s$ to optimize 49 models with the maximum degrees of $v(t)$ and $\dot{v}(t)$ ranging from one to seven, respectively. 
In the case of neural network models, 225.38 $s$ were required to optimize models with network depths up to 15.
Considering an average time for optimizing one model from each category, the neural network models required more time than the other models because of the many training epochs required to achieve an accurate prediction of $P(t)$.

The accuracy in the pressure prediction for each model category was also assessed.
For comparison, one benchmark model was selected from each model category that compromises between pressure estimation performance and model complexity.
The selected benchmark models were marked with $^*$ in Table~\ref{exp_results} and listed below:
\begin{enumerate}
    \item \textbf{exp}: A physics-inspired exponential model with three exponential terms of $v(t)$ ($k=3$)
    \item \textbf{NN}: A neural network model with eight fully connected layers ($d=8$)
    \item \textbf{NN-AR}: An autoregressive neural network model with an autoregressive order of five and eight fully connected layers ($p=5, d=8$)
    \item \textbf{poly}: A multivariate polynomial model with a maximum exponent of three for $v(t)$ and a maximum exponent of two for $\dot{v}(t)$ ($n=3, m=2$)
    \item \textbf{poly-AR}: An autoregressive multivariate polynomial model with an autoregressive order of five, a maximum exponent of three for $v(t)$, and a maximum exponent of two for $\dot{v}(t)$ ($p=5, n=3, m=2$)
\end{enumerate}

\begin{figure}[!t]
\centering
\includegraphics[width=0.48\textwidth]{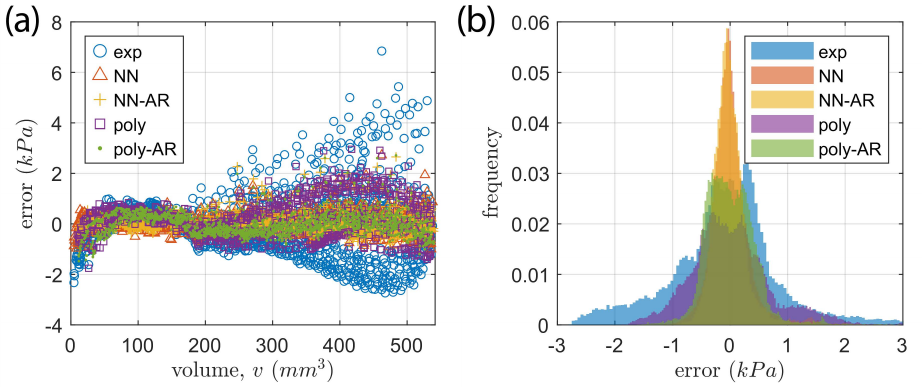}
\caption{
\textbf{True vs. estimated residual plots.}
(a)~Scatter plot of errors vs. volume, for the optimal models of each model type.
(b)~Error distributions for optimal models in each model type.
}
\label{fig_5}
\end{figure}

In Fig.~\ref{fig_5}(a), errors are scatter-plotted with respect to $v(t)$ for each benchmark model.
The physics-inspired exponential model exhibited greater errors in the larger volume regions with a maximum error of 6.84~$kPa$. 
The average and standard deviation of absolute errors were $0.76\pm0.81~kPa$.
For volumes greater than 300~$mm^3$, the hyper-elastic behavior has a greater influence than the fluid dynamics, and the model's limited number of parameters cannot capture this behavior.
In contrast, the neural network ($0.22\pm0.23\ kPa$) and multivariate polynomial models ($0.52\pm0.45\ kPa$) displayed a more even error distribution across volumes, with the neural network model generally producing lower errors.
In both cases, models that included autoregressive terms provided lower errors ($0.21\pm0.25\ kPa$ in NN\_AR and $0.33\pm0.28\ kPa$ in poly\_AR). 
This is presumably due to better handling of measurement noise and delays by using previous-state information.
In Fig.~\ref{fig_5}(b), error distribution plots for the benchmark models are shown.
The frequency of errors was obtained by generating histograms with a bin size of 0.05~$kPa$.
Notably, the errors from the neural network models were closer to a normal distribution, whereas the other models exhibited more model-specific error variations distributed across different regions.

\section{Practical Model Utilization Cases}
We demonstrate three practical use cases to illustrate how our model contributes to extending soft robots into dynamic regimes.
We focused on the benchmark multivariate polynomial model because its mathematical form made the forthcoming analyses convenient.
For example, the multivariate polynomial model in Table~\ref{Eqss} can be rewritten in the form below:
\begin{equation} \label{eom}
\begin{split}
    P(t) &= \sum_{i=0}^{n}\sum_{j=0}^m\alpha_{i,j}\bigl(v(t)\bigr)^i\bigl(\dot{v}(t)\bigr)^j\\
    &= \Bigl(\sum_{i=0}^{n}\sum_{j=1}^m\alpha_{i,j}\bigl(v(t)\bigr)^i\bigl(\dot{v}(t)\bigr)^{j-1}\Bigr)\dot{v}(t)+\sum_{i=0}^{n}\alpha_{i,0}\bigl(v(t)\bigr)^i \\
    &= C\bigl(v(t), \dot{v}(t)\bigr)\dot{v}(t)+K\bigl(v(t)\bigr),
\end{split}   
\end{equation}
which shows a clear connection to a standard equation of motion with negligible inertia.
Moreover, conventional mathematical procedures, such as taking partial derivatives with respect to $v(t)$ or $\dot{v}(t)$, can be easily performed since the equation explicitly includes polynomial (cross-)terms.  
Leveraging these features, we explore three direct applications.

\subsection{Stiffness/Damping Analysis with Air Pocket}
\begin{figure*}[!t]
\centering
\includegraphics[width=\textwidth]{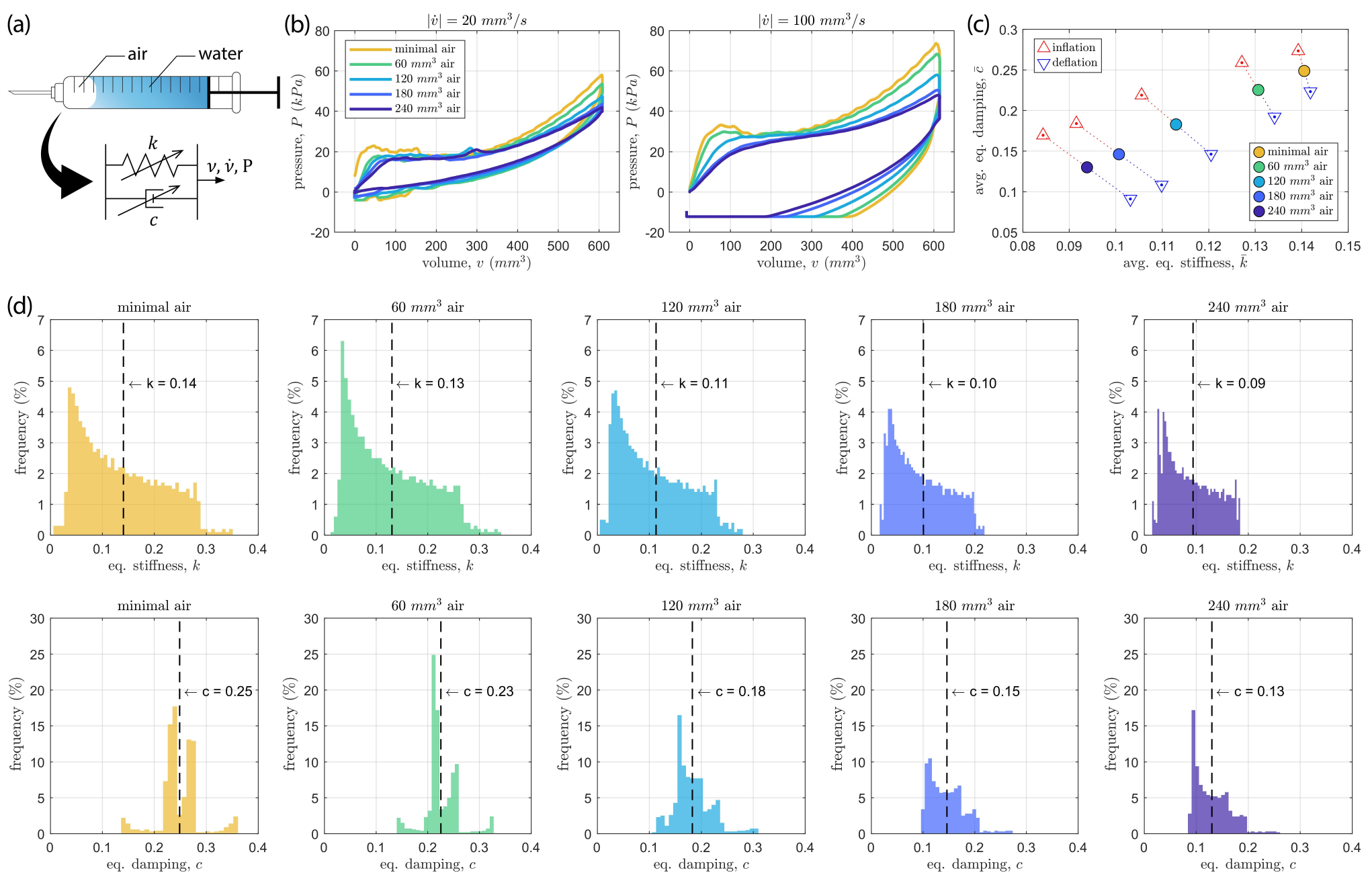}
\caption{
\textbf{Stiffness and damping analysis with different volume of air pockets.}
(a)~Schematic of an air pocket inside the actuation syringe. The air pocket functions as a spring-damper.
(b)~V--P plots for different volumes of air in a chamber across multiple inflation and deflation cycles. 
The left plot includes trials with $|\dot{v}| = 20\ mm^3$, and the right plot includes trials with $|\dot{v}| = 100\ mm^3$.
(c)~The plot displays the averaged equivalent stiffness ($\bar{k}$) and damping coefficients ($\bar{c}$) for different volumes of air in a chamber.
The red and blue triangles represent equivalent stiffness and damping coefficients achieved only across inflation and deflation cycles, respectively.
(d)~Distribution of pointwise $k_i$ and $c_i$ for different air volumes in a chamber. 
}
\label{fig_air}
\end{figure*}

Because of water’s incompressibility, hydraulic actuation offers precise control, high load-bearing capacity, and fast response.
However, maintaining a completely watertight system is difficult, and over time air infiltration can alter the actuator’s dynamic properties.
An air pocket within the hydraulic actuation tubing may reduce the stiffness and damping of actuation due to its compressibility (Fig.~\ref{fig_air}(a)).
If the presence of air pockets can be diagnosed and identified in real time, the model and control system can be optimally adjusted.

For a quantitative study, data were collected by injecting different amounts of air into the syringe using the same 3-DoF SBA from the modeling experiments, as shown in Fig.~\ref{fig_air}(b).
First, we computed partial derivatives of the multivariate polynomial model in Table~\ref{Eqss} with respect to $v(t)$ and $\dot{v}(t)$ at each data point as $k_i = \frac{\partial P}{\partial v}\big|_{(v(t_i),\dot{v}(t_i))}$ and $c_i = \frac{\partial P}{\partial \dot{v}}\big|_{(v(t_i),\dot{v}(t_i))}$, which we interpret as instantaneous stiffness $k_i$ and damping $c_i$.
%
Then, average stiffness, $\bar{k}$, and damping, $\bar{c}$, values were computed as $\bar{k} = \frac{1}{N}\sum_{i=1}^Nk_i$ and $\bar{c} = \frac{1}{N}\sum_{i=1}^Nc_i$.
%
These averaged stiffness and damping were also computed separately for the inflation and deflation phases.
Fig.~\ref{fig_air}(c) plots these average $k$ and $c$ for different air volumes, and Fig.~\ref{fig_air}(d) shows the distribution of pointwise $k_i$ and $c_i$.
Both plots show that larger air pockets result in lower overall stiffness and damping.
These results not only demonstrated that air infiltration can be detected and managed by observing the actuation-to-pressure relationship but also that system dynamics can be monitored in real time through high-resolution measurements of stiffness and damping.

\subsection{Identification of Different SBA Designs} \label{SYSID}
\begin{figure}[!t]
\centering
\includegraphics[width=0.48\textwidth]{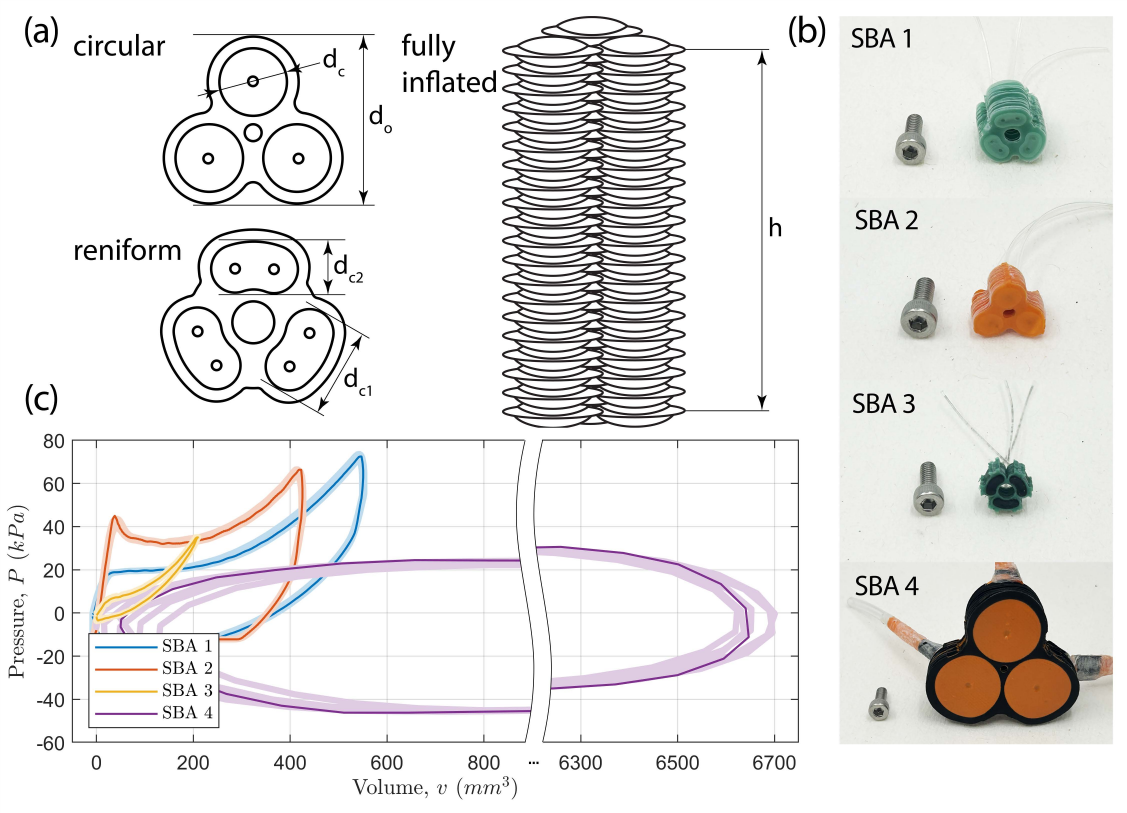}
\caption{
\textbf{Different characteristics of four SBAs.}
(a)~Geometric parameters of SBA.
(b)~Four different SBAs tested in this study. As a reference, M4$\times$10 $mm$ hex screw was placed next to each SBA.
(c)~V--P curves of the four different SBAs.
}
\label{fig_A2}
\end{figure}

SBAs can vary widely in geometry, materials, fabrication, and size, which may lead to distinct dynamic behaviors.
Characteristic SBA design parameters are depicted in Fig.~\ref{fig_A2}(a).
To assess volume-flow-pressure modeling performance across different SBAs, we fabricated four distinct 3-DoF SBAs (Fig.~\ref{fig_A2}(b)) and collected data for each to cross-validate the model.
The design parameters of the four SBAs are listed in Table~\ref{Diff_SBA}.
We collected data over multiple inflation/deflation cycles at various flow rates (Fig.~\ref{fig_A2}(c))
For validation, all 49 multivariate polynomial models with maximum exponents for $v(t)$ and $\dot{v}(t)$ from one to seven, respectively, were optimized on the dataset from each SBA.
%
The combination $n=3$ and $m=2$ exhibited the best performance in all four cases, with each model containing 12 coefficients.
Since all models share the same structure, averaged stiffness and damping values could be computed consistently across trajectories.
The results including evaluation performances and averaged stiffness/damping coefficients are shown in Table~\ref{Diff_SBA2}.

In addition to evaluating regression performance, the Chow test was applied to every combination of two models from different SBAs.
The Chow test assesses whether two regression models are statistically different~\cite{chow1960tests} or not.
%
The analysis involves computing F-statistics based on the ratio of covariances from an integrated model to those from two independent models.
Since all four SBAs were modeled using the same set of basis functions (e.g., the multivariate polynomial model with $n=3$ and $m=2$), each model is represented by its corresponding 12 coefficients.
Rather than simply differentiating the models using these coefficients, the Chow test was employed to quantitatively evaluate the ability to distinguish between datasets from different SBAs.
With a clear distinction between models from different SBAs, it becomes possible to identify a specific SBA even from a small dataset.
The results are shown in Table~\ref{Diff_SBA3}. 
All pairwise tests between models from SBA1 through SBA4 produced significant differences, with p-values less than $10^{-7}$, thereby rejecting the hypothesis that any two models are identical.
To further illustrate the significance, the F-statistics are listed in Table~\ref{Diff_SBA3}.
Since even the smallest F-statistic value ($F=1118.32$) was greater than the critical value of 5.9 at a significance level of $\alpha=0.0005$, the null hypothesis that any two models are identical was decisively rejected.

These results support the expanded applicability of the proposed volume-flow-pressure modeling approach, enabling cross-comparison among different SBAs using a single mathematical model form.
Moreover, the results suggest a link between the multivariate polynomial model with $n=3$ and $m=2$ and the underlying complex nonlinear physics governing SBA actuation.

\begin{table}[!t]
\caption{Design Parameters of Different SBAs}
\label{Diff_SBA}
\centering
\begin{tabular}{c|ccccccc}
\hline\hline
\multirow{2}*{No.} & chamber & \multirow{2}*{$d_o$} & $d_c$ & \multirow{2}*{$h$} & max. & max. & main \\
& shape & & ($d_{c1}, d_{c2}$) & & vol. & press. & material\\ \hline
1 & reniform & 14 & 7.4, 4.3 & 30 & 550 & 80 & TPE \\ 
2 & circular & 12 & 5.3 & 32 & 450 & 70 & TPU \\ 
3 & reniform  & 11.3 & 6.2, 2.4 & 20 & 200 & 35 & TPE \\ 
4 & circular  & 48 & 23.5 & 120 & 7000 & 40 & TPU \\ 
\hline
\multicolumn{2}{c}{} & $mm$ & $mm$ & $mm$ & $mm^3$ & $kPa$ & \\
\end{tabular}
\begin{itemize}    
    \item[] \footnotesize{$d_o$: outer width; $d_c$: chamber diameter ($d_{c1}$ and $d_{c2}$ in the reniform cases); $h$ maximum expandable height.
    }
\end{itemize}
\end{table}
 
\begin{table}[!t]
\caption{
Regression Performances for Different SBAs Using Multivariate Polynomial Model
}
\label{Diff_SBA2}
\centering
\begin{tabular}{c|ccccc}
\hline\hline
No. & RMSE & $R^2_{\mathrm{adj}}$ & AICc & $\bar{k}$ & $\bar{c}$ \\ \hline
1 & 1.4230 & 0.9943 & 3.5669e5 & 0.1561 & 0.1750 \\ 
2 & 3.8981 & 0.9783 & 1.5228e4 & 0.1774 & 0.2308 \\ 
3 & 0.5837 & 0.9974 & 2.0698e4 & 0.2305 & 0.5345 \\ 
4 & 7.1198 & 0.9440 & 7.4670e3 & 0.0003 & 0.9176 \\ 
\hline
\multicolumn{1}{c}{} & & & & $kPa/mm^3$ & $kPa\cdot s/mm^3$ \\
\end{tabular}
\end{table}

\begin{table}[!t]
\caption{Chow Test Results}
\label{Diff_SBA3}
\centering
\begin{tabular}{c|cccc}
 & SBA1 & SBA2 & SBA3 & SBA4 \\ \hline
SBA1 & - & 12301.55 & 6052.92 & 1118.32 \\
SBA2 & 12301.55 & - & 18891.99 & 6527.82 \\
SBA3 & 6052.92 & 18891.99 & - & 2477.83 \\
SBA4 & 1118.32 & 6527.82 & 2477.83 & - \\
\hline
\end{tabular}
\begin{itemize}    
    \item[] \footnotesize{The noted values are F-statistics, and the critical values for each case were 5.9 (at significance level of $\alpha=0.0005$).
    }
\end{itemize}
\end{table}

\subsection{External Force Estimation}
The models presented so far were based solely on intrinsic measurements from actuating SBAs, without any influence from external forces.
However, the pressure measurement will be affected by external forces.
A force sensor is typically required to decouple actuation-induced pressure from externally induced pressure, but sensing integration is challenging due to the SBA's form factor, material properties, and operating conditions.
Since the optimized models were proven to accurately predict the actuation-induced pressure, the difference between the estimated and measured pressures can be correlated to an external force. 
Based on this hypothesis, we proposed a sensorless (or ``proprioceptive'') external force estimation method. 
Assuming our model gives the true pressure in the absence of an external load, the pressure residual in each chamber is integrated to calculate the magnitude of the external force.

\begin{figure}[!t]
\centering
\includegraphics[width=0.48\textwidth]{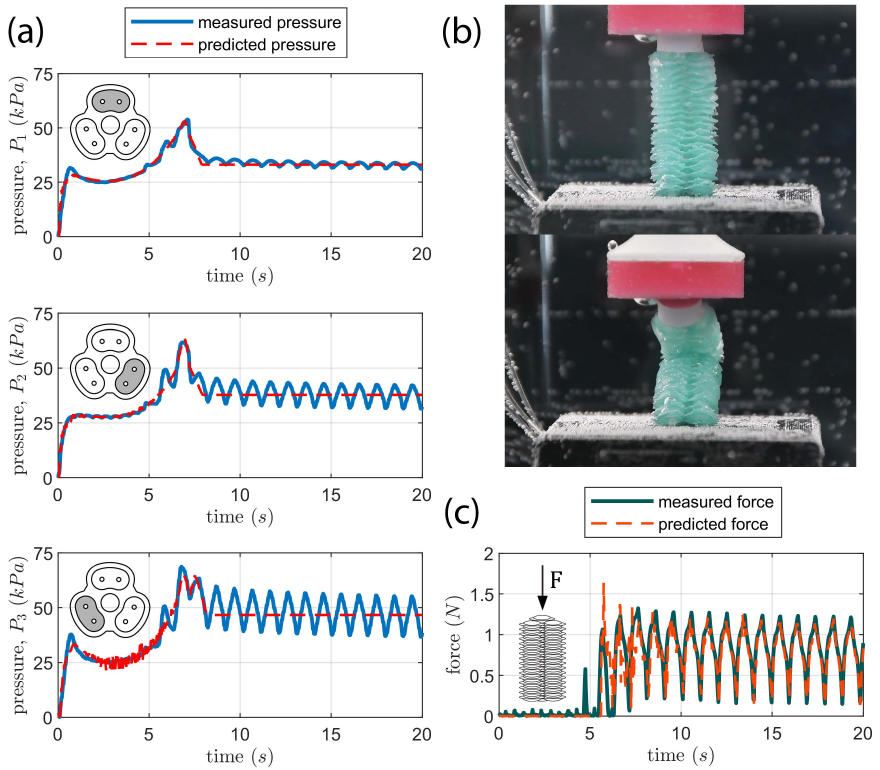}
\caption{
\textbf{Proprioceptive external force estimation based on pressure.}
(a)~Measured and estimated pressure in each of the three chamber.
(b)~Experiment for estimating the net external force applied to the SBA. 
(c)~The plot displays the measured and estimated net force applied to the SBA over time.
}
\label{fig_force}
\end{figure}

Using the procedure reported in Section~\ref{SYSID}, we identified a volume-flow-pressure model for each of the 3-DoF SBA’s three chambers. 
We then inflated the SBA with 500 $mm^3$ of water.
A robot arm (UR5e, UNIVERSAL ROBOTS) equipped with a force sensor (9105-TW-NANO17-E, ATI) was controlled to dynamically push the tip of the SBA 10~$mm$ repeatedly at a rate of 1~Hz.
The real-time pressure measurements and estimations based on individualized models for each chamber are shown in Fig.~\ref{fig_force}(a).
The force from the robot arm induced an asymmetric compression in the three chambers (Fig.~\ref{fig_force}(b)), leading to a different pressure in each of them.
The magnitude of the external force was calculated by multiplying the pressure differences by the effective cross-sectional area (22~$mm^2$ for each chamber) and summing the results.
%
The estimation performance was evaluated against the force sensor measurements, yielding a mean error of 1.60~$mN$ and a standard deviation of 7.50~$mN$.
Although only the magnitude of the force could be estimated, the method demonstrated high precision without the need for an additional sensor.

\section{Discussion and Conclusion}
Inflatable soft actuators, like most soft robots, exhibit complex, nonlinear behavior due to their inherent material properties.
Despite its advantages based on the incompressibility, hydraulic actuation adds complex dynamics throughout the actuation chain.
The inflatable soft actuators' behaviors become further complicated by the two practical issues frequently addressed in the field of soft robotics: fabrication inconsistencies (and/or defects) and difficulties in complete removal of air in the fluidic actuation chain.
These challenges inevitably necessitates continuous empirical tuning of the control system.
This is further exacerbated by the difficulty of integrating sensors for measuring such characteristics.

In this study, we introduced an approach to address these challenges by establishing a model that relates between the intrinsic measurements. 
The input fluidic volume ($v(t)$) and its time derivatives ($\dot{v}(t), \ddot{v}(t), ...$) as well as pressure ($P(t)$) are the most fundamental measurements that can be measured from the fluidic actuation chain without interfering the soft robotic behavior.

The actuation-to-pressure model was established by testing three types of models (e.g., physics-inspired exponential, neural network, and multivariate polynomial models) with incorporation of a model-selection criterion Eq.~(\ref{eq_optimization}) that balances between the prediction accuracy and model complexity.
The two key findings from the preliminary study were that second-(and-higher-)order time derivatives of volume impose negligible effect on pressure dynamics, whereas cross-terms between $v(t)$ and $\dot{v}(t)$ play a critical role.
Based on these features, optimal ($R^2 > 0.99$) yet simple ($\nu < 105$) models were achieved in each model type. 
The most notable outcome was that the multivariate polynomial model with a maximum exponent of three for $v(t)$ and a maximum exponent of two for $\dot{v}(t)$ showed optimal prediction performance compared to all the adjacent combinations, with only twelve coefficients.
This infers that the model is realizable and can be utilized in systems without intensive computation unit. 
Furthermore, as the model consists of explicit polynomials of $v(t)$ and $\dot{v}(t)$, this framework accepts conventional robotics techniques---from Kalman filtering to Koopman operator theory~\cite{bruder2020data}.
This intuition was shown with Eq.~\ref{eom}.

Since the optimal multivariate polynomial model ($n=3, m=2$) obtained with the benchmark SBA also yielded the highest pressure prediction performance across different SBAs, variations in SBA characteristics could be captured through system identification process while maintaining the same model structure. 
Owing to its simple form, real‑time, continuous model identification was feasible, as demonstrated in the air‑pocket diagnosis and SBA differentiation studies.
We envision that additional functionalities, such as fault detection and tracking time‑varying degradation, can be realized using a similar approach in a future work.
Furthermore, the model’s accurate pressure predictions enabled precise characterization of internal kinetics, thereby enabling precise proprioceptive force estimation without integrating a sensor.

In conclusion, we emphasize that the potential of this study is not limited to a specific form factor---SBA.
Since our models accommodate a range of nonlinearities such as hyper-elastic behavior, hysteresis effects, and measurement delays, they can be applied to elastomeric soft actuators and more complex soft robotic systems.
Building on this analytical framework, future work will focus on designing model‑based dynamic controllers.



\begin{thebibliography}{10}
\providecommand{\url}[1]{#1}
\csname url@samestyle\endcsname
\providecommand{\newblock}{\relax}
\providecommand{\bibinfo}[2]{#2}
\providecommand{\BIBentrySTDinterwordspacing}{\spaceskip=0pt\relax}
\providecommand{\BIBentryALTinterwordstretchfactor}{4}
\providecommand{\BIBentryALTinterwordspacing}{\spaceskip=\fontdimen2\font plus
\BIBentryALTinterwordstretchfactor\fontdimen3\font minus \fontdimen4\font\relax}
\providecommand{\BIBforeignlanguage}[2]{{%
\expandafter\ifx\csname l@#1\endcsname\relax
\typeout{** WARNING: IEEEtran.bst: No hyphenation pattern has been}%
\typeout{** loaded for the language `#1'. Using the pattern for}%
\typeout{** the default language instead.}%
\else
\language=\csname l@#1\endcsname
\fi
#2}}
\providecommand{\BIBdecl}{\relax}
\BIBdecl

\bibitem{rus2015design}
D.~Rus and M.~T. Tolley, ``Design, fabrication and control of soft robots,'' \emph{Nature}, vol. 521, no. 7553, pp. 467--475, 2015.

\bibitem{della2020model}
C.~Della~Santina, R.~K. Katzschmann, A.~Bicchi, and D.~Rus, ``Model-based dynamic feedback control of a planar soft robot: trajectory tracking and interaction with the environment,'' \emph{The International Journal of Robotics Research}, vol.~39, no.~4, pp. 490--513, 2020.

\bibitem{haggerty2023control}
D.~A. Haggerty, M.~J. Banks, E.~Kamenar, A.~B. Cao, P.~C. Curtis, I.~Mezi{\'c}, and E.~W. Hawkes, ``Control of soft robots with inertial dynamics,'' \emph{Science robotics}, vol.~8, no.~81, p. eadd6864, 2023.

\bibitem{armanini2023soft}
C.~Armanini, F.~Boyer, A.~T. Mathew, C.~Duriez, and F.~Renda, ``Soft robots modeling: A structured overview,'' \emph{IEEE Transactions on Robotics}, vol.~39, no.~3, pp. 1728--1748, 2023.

\bibitem{wang2018toward}
H.~Wang, M.~Totaro, and L.~Beccai, ``Toward perceptive soft robots: Progress and challenges,'' \emph{Advanced Science}, vol.~5, no.~9, p. 1800541, 2018.

\bibitem{lipson2014challenges}
H.~Lipson, ``Challenges and opportunities for design, simulation, and fabrication of soft robots,'' \emph{Soft Robotics}, vol.~1, no.~1, pp. 21--27, 2014.

\bibitem{polygerinos2017soft}
P.~Polygerinos, N.~Correll, S.~A. Morin, B.~Mosadegh, C.~D. Onal, K.~Petersen, M.~Cianchetti, M.~T. Tolley, and R.~F. Shepherd, ``Soft robotics: Review of fluid-driven intrinsically soft devices; manufacturing, sensing, control, and applications in human-robot interaction,'' \emph{Advanced engineering materials}, vol.~19, no.~12, p. 1700016, 2017.

\bibitem{cianchetti2014soft}
M.~Cianchetti, T.~Ranzani, G.~Gerboni, T.~Nanayakkara, K.~Althoefer, P.~Dasgupta, and A.~Menciassi, ``Soft robotics technologies to address shortcomings in today's minimally invasive surgery: the stiff-flop approach,'' \emph{Soft robotics}, vol.~1, no.~2, pp. 122--131, 2014.

\bibitem{niiyama2015pouch}
R.~Niiyama, X.~Sun, C.~Sung, B.~An, D.~Rus, and S.~Kim, ``Pouch motors: Printable soft actuators integrated with computational design,'' \emph{Soft Robotics}, vol.~2, no.~2, pp. 59--70, 2015.

\bibitem{SBA1}
T.~Ranzani, S.~Russo, F.~Schwab, C.~J. Walsh, and R.~J. Wood, ``Deployable stabilization mechanisms for endoscopic procedures,'' in \emph{2017 IEEE International Conference on Robotics and Automation (ICRA)}.\hskip 1em plus 0.5em minus 0.4em\relax IEEE, 2017, pp. 1125--1131.

\bibitem{SBAmed2}
J.~Rogatinsky, D.~Recco, J.~Feichtmeier, Y.~Kang, N.~Kneier, P.~Hammer, E.~O’Leary, D.~Mah, D.~Hoganson, N.~V. Vasilyev \emph{et~al.}, ``A multifunctional soft robot for cardiac interventions,'' \emph{Science Advances}, vol.~9, no.~43, p. eadi5559, 2023.

\bibitem{SBAlike2}
H.~D. Yang and A.~T. Asbeck, ``A layered manufacturing approach for soft and soft-rigid hybrid robots,'' \emph{Soft Robotics}, vol.~7, no.~2, pp. 218--232, 2020.

\bibitem{SBA3}
V.~T. Vo, L.~Z. Yañez, C.~Muter, A.~M. Moran, M.~Saxena, G.~Matthews, and T.~Ranzani, ``Soft, fiber-reinforced bellow actuators,'' \emph{IEEE Robotics and Automation Letters}, vol.~10, no.~2, pp. 1896--1903, 2025.

\bibitem{mosadegh2014pneumatic}
B.~Mosadegh, P.~Polygerinos, C.~Keplinger, S.~Wennstedt, R.~F. Shepherd, U.~Gupta, J.~Shim, K.~Bertoldi, C.~J. Walsh, and G.~M. Whitesides, ``Pneumatic networks for soft robotics that actuate rapidly,'' \emph{Advanced functional materials}, vol.~24, no.~15, pp. 2163--2170, 2014.

\bibitem{neohook2}
R.~W. Ogden, \emph{Non-linear elastic deformations}.\hskip 1em plus 0.5em minus 0.4em\relax Courier Corporation, 1997.

\bibitem{hagen1}
P.~Breitman, Y.~Matia, and A.~D. Gat, ``Fluid mechanics of pneumatic soft robots,'' \emph{Soft Robotics}, vol.~8, no.~5, pp. 519--530, 2021.

\bibitem{hurvich1993corrected}
C.~M. Hurvich and C.-L. Tsai, ``A corrected akaike information criterion for vector autoregressive model selection,'' \emph{Journal of time series analysis}, vol.~14, no.~3, pp. 271--279, 1993.

\bibitem{gaussian1}
J.~Vanhatalo, P.~Jyl{\"a}nki, and A.~Vehtari, ``Gaussian process regression with student-t likelihood,'' \emph{Advances in neural information processing systems}, vol.~22, 2009.

\bibitem{chow1960tests}
G.~C. Chow, ``Tests of equality between sets of coefficients in two linear regressions,'' \emph{Econometrica: Journal of the Econometric Society}, pp. 591--605, 1960.

\bibitem{bruder2020data}
D.~Bruder, X.~Fu, R.~B. Gillespie, C.~D. Remy, and R.~Vasudevan, ``Data-driven control of soft robots using koopman operator theory,'' \emph{IEEE Transactions on Robotics}, vol.~37, no.~3, pp. 948--961, 2020.

\end{thebibliography}
\end{document}